%% file: main.tex
\documentclass[runningheads]{llncs}

\usepackage{eccv}

\usepackage{eccvabbrv}

\usepackage{graphicx}
\usepackage{booktabs}
\usepackage{multirow}

\usepackage[accsupp]{axessibility}  

\usepackage{hyperref}

\usepackage{orcidlink}
\newcommand{\model}{DualCount\xspace}

\begin{document}

\title{DualCount: Structurally Consistent Density and Point Modeling for Zero-Shot Object Counting} 

\titlerunning{DualCount}

\author{Xuan Cuong Ngo\orcidlink{0009-0003-3938-3678}}

\authorrunning{Ngo}

\institute{University of Arkansas, USA\\
\email{cngo@uark.edu}}

\maketitle
\input{section/abstract}
\input{section/introduction}
\input{section/related_work}
\input{section/method}
\input{section/experiment}

\input{section/conclusion}

\bibliographystyle{splncs04}
\bibliography{main}
\end{document}

%% file: section/abstract.tex
\begin{abstract}
Zero-shot object counting aims to estimate the number of objects specified by a text query without category-specific training. Recent approaches primarily rely on density regression or detection-style instance prediction. While effective, density-based models often suffer from spatial ambiguity and background leakage due to weakly regulated mass allocation, leading to fragmented or part-biased representations that increase counting error in complex scenes.
In this work, we propose an instance-aware dual-decoder framework that structurally couples density and point representations for zero-shot object counting. Instead of treating density estimation as independent pixel-wise regression, we interpret it as a structured mass allocation problem over a latent set of object instances. Predicted instance centers induce a soft instance-wise decomposition of the density map, upon which we enforce two geometric constraints: (1) per-instance mass conservation, ensuring each object contributes approximately one unit of density mass, and (2) center-of-mass alignment, encouraging each density component to concentrate around its corresponding predicted center. These constraints introduce instance-level geometric consistency and lead to more accurate mass allocation, thereby reducing counting error.
Extensive experiments on FSC-147, PUCPR+, and CARPK show that our approach consistently reduces counting error and establishes new state-of-the-art performance in zero-shot object counting.

  \keywords{Object Counting \and Point Prediction \and Zero-shot Learning}
\end{abstract}

%% file: section/introduction.tex
\section{Introduction}
\label{sec:intro}
Object counting traditionally focuses on estimating the number of instances belonging to a predefined class within an image, such as crowd counting \cite{sindagi2018survey}, vehicle counting \cite{guerrero2015extremely}, or animal counting \cite{qian2023counting}. These approaches rely on category-specific supervision and struggle to generalize beyond trained object categories. To improve generalization, class-agnostic few-shot counting has been introduced, where the target object is specified at inference time using visual exemplars, typically user-provided bounding boxes highlighting a few instances. Although such methods enable open-world counting, they require manual intervention at test time, limiting full automation and scalability.

\input{section/teaser}
To address this limitation, recent work explores zero-shot object counting, where the object of interest is specified using a textual description instead of visual exemplars \cite{xu2023zero,xc_bmvc,xc_soict}. By leveraging vision–language models, these approaches allow arbitrary object categories to be queried at inference time using class names. Most existing zero-shot counting methods adopt density regression, estimating object count by integrating a predicted density map conditioned on text prompts. 
While effective, this formulation has a key limitation: density supervision provides limited guidance on instance-level spatial structure. Although the predicted density map is encouraged to match the target density at each location, it does not explicitly regulate how the density mass should be distributed across individual object instances.

This limited instance-level guidance leads to spatial ambiguity. In practice, density mass may accumulate on correlated background regions or concentrate on highly discriminative object parts rather than stable instance centers. For objects with complex shapes, heavy occlusion, or cluttered surroundings, such ambiguous mass allocation results in fragmented instance representations and increased counting error. This failure case can be observed in Fig.~\ref{fig:teaser}.
Importantly, this issue arises not from weak semantic alignment alone, but from the absence of instance-level structural constraints within the density formulation itself.

Our key insight is that density maps implicitly represent a mixture of object-level components, yet existing approaches treat them as unconstrained continuous fields. If each object contributes approximately one unit of mass, then density estimation can be reformulated as a structured mass allocation problem over a latent set of object instances. Enforcing geometric consistency at the instance level can therefore directly improve counting reliability.

Based on this insight, we propose \textbf{DualCount}, an instance-aware dual-decoder framework for zero-shot object counting. Our model jointly predicts a density map and a set of instance centers from shared visual–text features. Rather than treating these predictions independently, we interpret them as complementary views of a shared latent object set. The predicted centers induce a soft instance-wise decomposition of the density map, allowing each object to be represented as an individual density component.

On top of this decomposition, we introduce two geometric constraints. First, a per-instance mass conservation constraint ensures that each object contributes approximately one unit of density mass, directly aligning spatial allocation with the counting objective. Second, a center-of-mass alignment constraint encourages each density component to concentrate around its corresponding predicted center, promoting coherent and compact instance representations. Together, these constraints transform density estimation from pure pixel-wise regression into a structured instance-level modeling problem. By reducing ambiguous mass allocation and preventing part-biased density drift, our approach leads to more accurate mass distribution and consequently reduces counting error.

Extensive experiments on FSC-147, PUCPR+, and CARPK demonstrate that our method consistently reduces counting error and establishes new state-of-the-art performance in zero-shot object counting. The improvements are particularly pronounced in crowded and geometrically complex scenes, where standard density regression tends to produce unstable spatial representations.

The contributions of this paper are threefold:

\begin{itemize}
    \item \textbf{Instance-aware formulation for zero-shot counting.} We reinterpret density estimation as a structured mass allocation problem over a latent set of object instances, revealing the importance of instance-level geometric constraints.
    
    \item \textbf{Geometric consistency through dual decoders.} We propose an instance-aware dual-decoder framework that couples density and point representations via per-instance mass conservation and center-of-mass alignment constraints.

    \item \textbf{Improved counting performance.} Extensive experiments and ablation studies on FSC-147, PUCPR+, and CARPK demonstrate consistent reductions in counting error and new state-of-the-art results in zero-shot object counting.
\end{itemize}

%% file: section/teaser.tex
\begin{figure*}[t] 
    \centering
    \includegraphics[width=\linewidth]{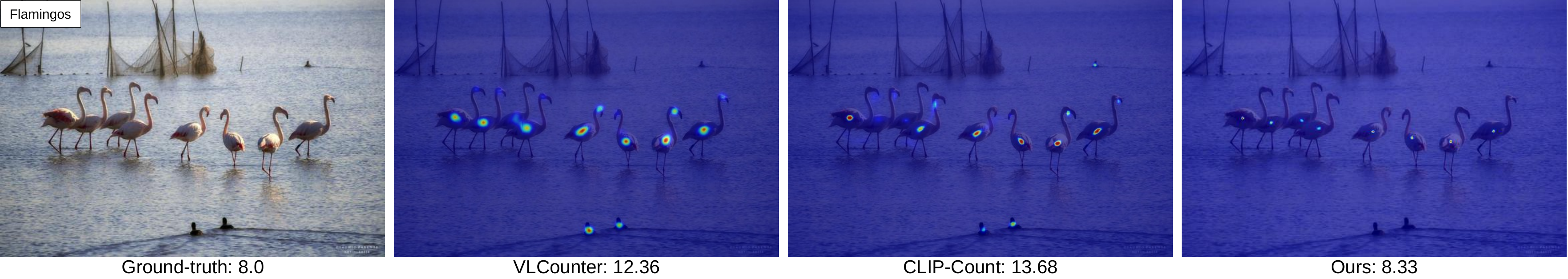}
    \caption{Density map visualizations. CLIP-Count \cite{jiang2023clip} and VLCounter \cite{kang2024vlcounter} are prone to false positive predictions, while our model mitigates this issue by leveraging spatial constraints from the point prediction branch, resulting in more controlled density estimates.}
    \label{fig:teaser}
\end{figure*}

%% file: section/related_work.tex
\section{Related Work}
\label{sec:related}
\textbf{Class-Specific Object Counting.} 
Object counting has long been studied in computer vision, with early research primarily focusing on specific object categories such as cars, cells, pedestrians, or polyps. Detection-based approaches were widely adopted in this stage, where object detectors are trained to localize and count instances of predefined categories. Modern detection frameworks such as YOLO and Faster R-CNN have been successfully applied in vehicle counting and surveillance systems, while segmentation-based methods are often used in biological cell counting \cite{tyagi2023degpr} and crowd analysis \cite{wang2020distribution}. 
However, detection-based counting typically struggles in highly crowded scenes due to severe occlusions and object overlaps. To alleviate these limitations, density-based methods were introduced, which estimate object counts by regressing a continuous density map whose integral corresponds to the object count. Density-based formulations are more robust to occlusions and scale variations, and have become the dominant paradigm in crowd counting and dense object counting scenarios. Subsequent studies further improved density-based supervision beyond standard Gaussian-map regression.
Bayesian Loss~\cite{ma2019bayesian} models probabilistic pixel-to-point contributions, while DM-Count~\cite{wang2020distribution} formulates counting as distribution matching with an optimal transport (OT)-based loss. Our density decoder follows this OT-based formulation, but further introduces instance-level geometric constraints to guide density allocation around individual objects.

\textbf{Class-Agnostic Object Counting.} 
To improve generalization beyond predefined categories, class-agnostic few-shot counting has recently gained significant attention. In this setting, the target object is specified at inference time using a small number of visual exemplars, typically provided as bounding boxes. Early methods adopt Siamese matching networks to learn similarity between exemplars and image regions for density estimation. BMNet+ \cite{min2022bmnet} incorporates non-linear similarity metrics for improved matching, while FamNet \cite{ranjan2021learning} enhances backbone design for better density prediction. CounTR \cite{liu2022countr} combines convolutional feature extraction with transformer-based modeling and cross-attention to improve feature fusion. SAFECount \cite{you2023few} introduces feature refinement modules to enhance generalization, and LOCA \cite{djukic2023low} proposes object prototype extraction for iterative adaptation to exemplar appearance and shape.
Despite their effectiveness, these approaches depend heavily on the quality and representativeness of visual exemplars. Sparse, biased, or noisy exemplars can degrade performance. Moreover, requiring manual exemplar input at inference time limits scalability and full automation in open-world scenarios.

\textbf{Text-Specified Object Counting.} 
Text-specified or zero-shot object counting eliminates the need for visual exemplars by allowing users to specify the target object using textual descriptions. Leveraging vision–language models, these approaches enable open-world counting across arbitrary object categories. Paiss et al.~\cite{paiss2023teaching} demonstrated that fine-tuned CLIP models can perform limited counting. Building upon this direction, CLIP-Count \cite{jiang2023clip} and VLCounter \cite{kang2024vlcounter} condition density regression on text embeddings extracted from CLIP, enabling open-vocabulary density-based counting. TFOC \cite{shi2023trainingfreeobjectcountingprompts} further explores training-free counting by prompting the segmentation model SAM \cite{kirillov2023segany} with text, points, or boxes.
While these methods demonstrate promising generalization, most existing approaches rely on density regression with limited instance-level spatial guidance. As a result, although the predicted density map is encouraged to match the target density, supervision does not explicitly regulate how density mass should be distributed across object instances.
This can lead to background leakage, fragmented instance representations, or part-biased density allocation, particularly in cluttered or geometrically complex scenes.
In contrast, our work introduces instance-level geometric consistency into density-based zero-shot counting. By coupling density estimation with predicted instance centers and enforcing per-instance mass conservation and center-of-mass alignment constraints, we transform density regression into a structured mass allocation problem, reducing ambiguous spatial distribution and improving counting reliability.

%% file: section/method.tex
\section{Methodology}
\label{sec:method}

Given an input image $I \in \mathbb{R}^{H \times W \times 3}$ and a text query $t$ specifying the target object category, zero-shot object counting aims to estimate the number of instances of the queried category in the image. Let $N$ denote the ground-truth object count and $\hat{N}$ the predicted count. In this setting, the object category is defined solely by the textual query $t$, and the model is required to predict the corresponding object count.

\input{section/model}

\subsection{Architecture Overview.}
Figure \ref{fig:model} presents the proposed \model framework. Our framework consists of five components: (1) an image encoder $E_v$, (2) a text encoder $E_t$, (3) a cross-modal feature enhancement module $\mathcal{E}$, (4) a feature fusion module $\mathcal{F}$, and (5) a dual-decoder head composed of a density decoder $D_{\rho}$ and a point decoder $D_c$. The image and text encoders extract unimodal representations, which are subsequently refined and fused into a text-conditioned spatial feature map. The dual decoders then produce complementary outputs, a density map for count estimation and a set of instance centers for instance-level structural modeling.

\subsubsection{Encoders.}
We employ a visual encoder $E_v$ and a text encoder $E_t$ to extract unimodal representations from the input image and text query. The image encoder maps $I \in \mathbb{R}^{H \times W \times 3}$ to a spatial feature map $F_v = E_v(I) \in \mathbb{R}^{C \times H' \times W'}$, where $C$ denotes the feature dimension and $(H', W')$ are the spatial resolutions after downsampling. Each spatial location corresponds to a $C$-dimensional feature vector encoding local visual information. The text encoder projects the query $t$ into a semantic embedding $f_t = E_t(t) \in \mathbb{R}^{C}$. The visual and textual features share the same embedding dimension, enabling effective cross-modal interaction in subsequent modules.

\subsubsection{Feature Enhancement via Cross-Modal Attention.}
Although the backbone feature map $F_v \in \mathbb{R}^{C \times H' \times W'}$ encodes rich visual semantics, it is not explicitly optimized for fine-grained object counting. Counting requires both precise spatial discrimination and strong semantic alignment with the queried object. To strengthen the representation, we design a cross-modal enhancement module $\mathcal{E}(\cdot)$ composed of image self-attention and bidirectional cross-attention.
We first reshape $F_v$ into a sequence of spatial tokens $\mathbf{V} \in \mathbb{R}^{(H'W') \times C}$, where each row corresponds to a spatial location. 

\noindent
\textbf{Image Self-Attention.}
We apply multi-head self-attention to capture long-range spatial dependencies among image features. 
Given the visual feature sequence $\mathbf{V}$, the query, key, and value features are computed as 
$\mathbf{Q}_v = \mathbf{V}W_q^v$, 
$\mathbf{K}_v = \mathbf{V}W_k^v$, and 
$\mathbf{V}_v = \mathbf{V}W_v^v$, 
where $W_q^v, W_k^v, W_v^v \in \mathbb{R}^{C \times C}$ are learnable projection matrices. 
The self-attended features are then obtained as
\begin{equation}
\mathbf{V}^{\mathrm{sa}} 
= 
\operatorname{Softmax}\!\left(
(\mathbf{Q}_v \mathbf{K}_v^\top)/\sqrt{C}
\right)
\mathbf{V}_v .
\end{equation}
This operation allows each spatial location to aggregate contextual information from all other regions, producing more robust visual features for counting.

\noindent
\textbf{Image-to-Text Cross-Attention.}
To incorporate semantic guidance from the queried category, we allow visual tokens to attend to the text embedding $f_t \in \mathbb{R}^{C}$. 
The text key and value are computed as 
$\mathbf{k}_t = f_t W_k^t$ and 
$\mathbf{v}_t = f_t W_v^t$, 
while the visual queries are computed as 
$\mathbf{Q}_{vt} = \mathbf{V}^{\mathrm{sa}} W_q^{vt}$. 
The cross-attended visual features are then obtained by
\begin{equation}
\mathbf{V}^{vt}
=
\operatorname{Softmax}\!\left(
(\mathbf{Q}_{vt}\mathbf{k}_t^\top)/\sqrt{C}
\right)
\mathbf{v}_t .
\end{equation}
This step conditions the visual representation on the queried category and highlights spatial regions that are semantically relevant for counting.



\noindent
\textbf{Text-to-Image Cross-Attention.}
To further strengthen alignment, the text representation attends back to the visual features. 
We compute $\mathbf{q}_t = f_t W_q^{tv}$, 
$\mathbf{K}_{tv} = \mathbf{V}^{\mathrm{sa}} W_k^{tv}$, and 
$\mathbf{V}_{tv} = \mathbf{V}^{\mathrm{sa}} W_v^{tv}$. 

\noindent
The updated text feature is computed as
\begin{equation}
f_t'
=
\operatorname{Softmax}\!\left(
(\mathbf{q}_t \mathbf{K}_{tv}^{\top})/\sqrt{C}
\right)
\mathbf{V}_{tv},
\end{equation}
which encourages tighter cross-modal correspondence.

\noindent
\textbf{Feed-Forward Refinement.}
Finally, the enhanced visual tokens and text embedding are refined via position-wise feed-forward networks, yielding $\mathbf{V}^{enh} = \text{FFN}_v(\mathbf{V}^{vt})$ and $f_t^{enh} = \text{FFN}_t(f_t')$. The visual tokens are reshaped back to spatial form $\tilde{F}_v \in \mathbb{R}^{C \times H' \times W'}$.

The enhanced feature map $\tilde{F}_v$ encodes strengthened spatial context and improved semantic alignment with the text query, providing a robust foundation for both density estimation and center prediction.

\subsubsection{Feature Fusion.}

After cross-modal enhancement, we obtain the refined visual feature map $\tilde{F}_v \in \mathbb{R}^{C \times H' \times W'}$ and the enhanced text embedding $f_t^{enh} \in \mathbb{R}^{C}$. We fuse these modalities to construct a text-conditioned spatial feature map $F$ that serves as input to the dual decoders.

We adopt feature-wise modulation to inject global semantic information from the text embedding into spatial visual features. Specifically, we generate channel-wise modulation parameters from the text feature:
\begin{equation}
    \gamma = W_\gamma f_t^{enh}, \quad
    \beta = W_\beta f_t^{enh},
\end{equation}
where $W_\gamma, W_\beta \in \mathbb{R}^{C \times C}$ are learnable projections, and $\gamma, \beta \in \mathbb{R}^{C}$.

The fused feature map is computed as
\begin{equation}
    F(x) = \gamma \odot \tilde{F}_v(x) + \beta,
\end{equation}
for each spatial location $x \in \Omega$, where $\Omega$ denotes the set of all spatial coordinates in the image domain and $\odot$ denotes channel-wise multiplication.

This modulation mechanism allows the text embedding to dynamically reweight and shift visual feature channels, emphasizing dimensions relevant to the queried object category while suppressing irrelevant responses. As a result, the fused feature map $F \in \mathbb{R}^{C \times H' \times W'}$ encodes spatially grounded representations that are conditioned on the target semantics. This representation is then shared by both the density decoder and the center decoder.

\subsubsection{Dual Decoder}
\label{sec:dual}
Given the fused feature map $F \in \mathbb{R}^{C \times H' \times W'}$, we design a dual-decoder architecture that produces (i) a density map for count estimation and (ii) a binary point map for instance-level supervision.

\noindent
\textbf{Shared Upsampling and Refinement.}
Both decoders share a lightweight refinement head composed of bilinear interpolation layers for progressive upsampling and $2\times2$ convolutional layers for channel reduction. Let $F^{up}$ denote the upsampled and refined feature map at the original image resolution $(H, W)$. This shared structure ensures that both density and point predictions are derived from spatially aligned high-resolution features.

\noindent
\textbf{Density Prediction Head.}
The density prediction head applies a fully connected feed-forward network (FFN) on $F^{up}$ to produce a scalar density value at each spatial location:
$\rho = \text{FFN}_{\rho}(F^{up}) \text{ with } \rho \in \mathbb{R}^{H \times W}.$
The predicted count is obtained by integrating the density map: $\hat{N} = \sum_{x \in \Omega} \rho(x).$
Instead of pixel-wise regression, following \cite{xc_soict} we supervise the density map using an optimal transport (OT) loss $\mathcal{L}_{ot}$ with the ground-truth point annotations. Let $\mathcal{P}^{gt} = \{p_j\}_{j=1}^{N}$ denote the set of ground-truth object centers. The OT loss aligns the predicted density distribution $\rho$ with the discrete empirical distribution defined by $\mathcal{P}^{gt}$, encouraging mass to concentrate around true object locations while preserving global count consistency. This formulation provides spatially meaningful supervision beyond simple $\ell_2$ density regression.

\noindent
\textbf{Point Prediction Head.}
In parallel with density estimation, we predict a sparse set of candidate instance centers using a point prediction branch. Concretely, the point head outputs a \emph{logit} map $S = \mathrm{FFN}{c}(F^{up}) \in \mathbb{R}^{H\times W},$
which is converted to a confidence map by a sigmoid, $h(x)=\sigma(S(x))\in(0,1)$ with $x\in\Omega.$
At training and inference time, we extract a set of center proposals
$\hat{\mathcal P}=\{\hat p_i\}_{i=1}^{M}$
by first applying non-maximum suppression (NMS) to the confidence map $h$
with a fixed window size, and then selecting all remaining local maxima whose
confidence exceeds a threshold $\tau$:
\begin{equation}
\hat{\mathcal P}
=
\left\{
x \in \Omega \;\middle|\;
x \text{ is a local maximum under NMS and } h(x)\ge \tau
\right\}.
\end{equation}
Each proposal $\hat p_i=(u_i,v_i)$ is associated with a confidence score
$s_i=h(\hat p_i)$, and the number of proposals $M=|\hat{\mathcal P}|$ varies per image. To train this, we use Focal loss $\mathcal{L}_{focal}$ following \cite{NEURIPS2024_57c56985}.
To further establish a one-to-one correspondence between predicted proposals and ground-truth centers $\mathcal{P}^{gt}=\{{p_j}\}_{j=1}^{N}$, we solve a bipartite matching problem using the Hungarian algorithm. The matching cost is quadric cost
and we compute an optimal assignment $\mathcal M \subset \hat{\mathcal P}\times\mathcal P^{gt}$. Unmatched predictions are treated as negatives (false positives), and unmatched ground-truth points are treated as missed detections.
For each predicted point $\hat p_i$, we first feed the input image $I$ into a CNN backbone to obtain multi-scale feature maps. We then extract a discriminative point-wise feature for $\hat p_i$ by bilinear sampling from these feature maps.
Let $\{{F^{l_j}}\}_{j=1}^{L}$ denote features from $L$ selected layers (e.g., intermediate decoder stages). For each scale $l_j$, we sample at location $\hat p_i$ using bilinear interpolation $f^{l_j}(\hat p_i) = \mathrm{Bilinear}\!\left(F^{l_j}, \hat p_i\right).$
We then concatenate features across scales to obtain the final point representation:
\begin{equation}
f(\hat p_i)=\mathrm{Concat}\left(f^{l_1}(\hat p_i),\dots,f^{l_L}(\hat p_i)\right)\in\mathbb{R}^{d}.
\end{equation}
Following \cite{xc_bmvc}, we apply a contrastive objective to encourage matched predictions to be similar to their corresponding ground-truth counterparts while pushing away unmatched predictions. Specifically, for each matched pair $(\hat p_i, p_j)\in\mathcal M$, we treat $f(\hat p_i)$ and $f(p_j)$ as a positive pair, while features extracted at unmatched predicted points $\hat p_k$ (or randomly sampled background locations) serve as negatives. We use an InfoNCE-style loss:
\begin{equation}
\mathcal L_{\mathrm{cst}}
=
-\frac{1}{|\mathcal M|}
\sum_{(\hat p_i,p_j)\in\mathcal M}
\log
\frac{
\exp\!\big(\mathrm{sim}(f(\hat p_i),f(p_j))/\tau\big)
}{
\sum\limits_{q\in \{p_j\}\cup \mathcal N_i}
\exp\!\big(\mathrm{sim}(f(\hat p_i),f(q))/\tau\big)
}.
\end{equation}
where $\mathrm{sim}(\cdot,\cdot)$ denotes cosine similarity, $\tau$ is a temperature hyperparameter, and $\mathcal N_i$ is the negative set (unmatched predictions and/or background samples). This loss improves the discriminability of point features and suppresses false-positive centers by contrasting them against matched instances.

\subsection{Soft Instance Decomposition}

The density map $\rho \in \mathbb{R}^{H \times W}$ provides a continuous representation for counting but does not explicitly encode instance-level structure. To introduce object-wise modeling, we decompose $\rho$ into soft instance-specific components guided by the predicted point set $\hat{\mathcal{P}} = \{\hat{p}_i\}_{i=1}^{M}$ obtained from the point decoder, where each $\hat{p}_i = (u_i, v_i)$ denotes a 2D coordinate on the image grid.

For each spatial location $x \in \Omega$, we define a soft assignment weight with respect to each predicted point using a Gaussian kernel:
\begin{equation}
w_i(x) =
\frac{\exp\left(-\frac{\|x - \hat{p}_i\|^2}{2\sigma^2}\right)}
{\sum_{j=1}^{M} \exp\left(-\frac{\|x - \hat{p}_j\|^2}{2\sigma^2}\right)},
\end{equation}
where $\sigma$ controls the spatial spread of the assignment and $\sum_{i=1}^{M} w_i(x) = 1$ for all $x \in \Omega$.
The global density map is then decomposed as
\begin{equation}
\rho_i(x) = w_i(x)\rho(x),
\end{equation}
which satisfies $\rho(x) = \sum_{i=1}^{M} \rho_i(x).$

Each component $\rho_i$ represents the soft density contribution associated with the predicted point $\hat{p}_i$. This decomposition introduces differentiable instance-level structure into the density representation.

\subsection{Per-Instance Mass Conservation}
\label{sec:mass}
Although standard density-based counting supervises the predicted density map with pixel-wise regression, it provides limited explicit control over how much density mass is assigned to each individual instance. To address this limitation, we introduce Per-Instance Mass Conservation, which constrains each object instance to receive a consistent unit of density mass.

Given the decomposition $\{\rho_i\}_{i=1}^{M}$, we compute the total mass assigned to instance $i$ as
\begin{equation}
m_i = \sum_{x \in \Omega} \rho_i(x).
\end{equation}

Ideally, each object contributes approximately one unit of density mass, i.e., $m_i \approx 1$ for valid instances. We therefore introduce a per-instance mass conservation loss:
\begin{equation}
\mathcal{L}_{\text{mass}} =
\frac{1}{M}\sum_{i=1}^{M} (m_i - 1)^2.
\end{equation}

This regularization encourages balanced mass allocation across predicted instances and prevents excessive concentration or dispersion of density mass.

\subsection{Center-of-Mass Alignment}
\label{sec:align}
Beyond mass conservation, we further enforce geometric consistency between density allocation and predicted points. For each density component $\rho_i$, we compute its centroid as
\begin{equation}
\bar{p}_i =
\frac{1}{m_i} \sum_{x \in \Omega} x \, \rho_i(x),
\end{equation}
where $x = (u, v)$ denotes spatial coordinates and $m_i$ is defined above.

We then align the centroid $\bar{p}_i$ with the corresponding predicted point $\hat{p}_i$ using
\begin{equation}
\mathcal{L}_{\text{align}} =
\frac{1}{M}\sum_{i=1}^{M}
\| \bar{p}_i - \hat{p}_i \|_2^2.
\end{equation}

This alignment encourages each density component to concentrate around its associated predicted point, reducing fragmented or part-biased mass allocation and promoting instance-level geometric coherence.

\subsection{Overall Loss Function}

In addition to $\mathcal{L}_{ot}$, $\mathcal{L}_{cst}$, $\mathcal{L}_{focal}$, $\mathcal{L}_{mass}$, and $\mathcal{L}_{align}$ introduced in previous sections, we enforce agreement between the two counting mechanisms through a count consistency loss that penalizes discrepancies between the count estimated from the density map and the number of predicted points:
\begin{equation}
\mathcal{L}_{\text{consistent}} =
\left(\hat{N} - M\right)^2,
\end{equation}
where $\hat{N}=\sum_{x\in\Omega}\rho(x)$ denotes the total mass of the predicted density map and $M$ is the number of predicted points.

The final training objective is

\begin{equation}
\mathcal{L} =
\lambda_{\text{ot}}\mathcal{L}_{\text{ot}}
+ \lambda_{\text{cst}}\mathcal{L}_{\text{cst}}
+ \lambda_{\text{focal}}\mathcal{L}_{\text{focal}}
+ \lambda_{\text{consistent}}\mathcal{L}_{\text{consistent}} \\
+ \lambda_{\text{mass}}\mathcal{L}_{\text{mass}}
+ \lambda_{\text{align}}\mathcal{L}_{\text{align}},
\label{eq:loss}
\end{equation}

where $\lambda_{\text{ot}}, \lambda_{\text{cst}}, \lambda_{\text{focal}}, \lambda_{\text{consistent}}, \lambda_{\text{mass}}$, and $\lambda_{\text{align}}$ balance the contributions of the loss terms.

%% file: section/model.tex
\begin{figure*}[t]
    \centering
    \includegraphics[width=\linewidth]{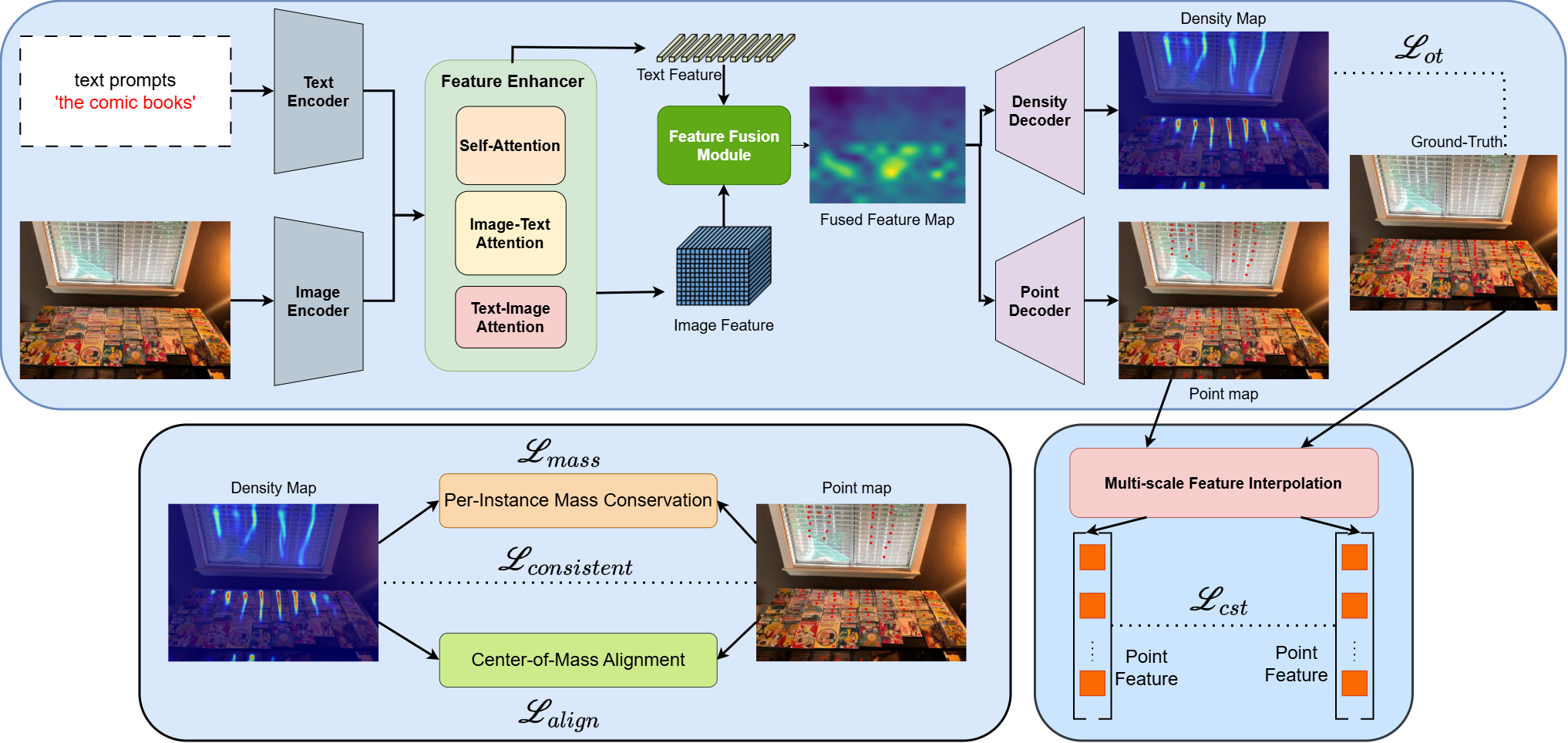}
    \caption{Overview of DualCount. Visual and textual features are extracted by the image encoder $E_v$ and text encoder $E_t$, enhanced via cross-modal interaction $\mathcal{E}$, and fused by $\mathcal{F}$ to produce a fused feature map. A dual-decoder head ($D_{\rho}$ and $D_c$) predicts the density map and instance centers.}
    \label{fig:model}
\end{figure*}

%% file: section/experiment.tex
\section{Experiments}
\subsection{Datasets and Evaluation Metrics}

\input{section/qual}
We conduct experiments on three widely used object counting benchmarks: FSC-147, CARPK, and PUCPR+. 

\textbf{FSC-147}~\cite{ranjan2021learning} is a large-scale few-shot counting dataset comprising 6,135 images across diverse object categories. The number of instances per image ranges from 7 to 3,731, with an average of 56. Each image is paired with three exemplar crops, randomly chosen object instances annotated with bounding boxes. The dataset is split into 89 categories for training, and 29 disjoint categories each for validation and testing, making FSC-147 a challenging open-set benchmark.  

\textbf{CARPK}~\cite{hsieh2017drone} focuses on vehicle counting in aerial imagery, captured by drones over parking lots. It provides 989 images for training and 459 for testing, with annotations for more than 89,000 cars.  

\textbf{PUCPR+}~\cite{hsieh2017drone} is another vehicle counting dataset collected from fixed surveillance cameras at the Pontifical Catholic University of Paraná, serving as a complementary fixed-camera benchmark to the drone-based CARPK dataset.

\medskip
\noindent
We assess model performance using two standard evaluation metrics: Mean Absolute Error (MAE) and Root Mean Squared Error (RMSE):  

\begin{equation}
\resizebox{\columnwidth}{!}{$
\begin{alignedat}{2}
MAE  &= \frac{1}{N_I}\sum_{i=1}^{N_{I}}\lvert N^{pred}_i - N^{gt}_i \rvert 
&\qquad
RMSE &= \sqrt{\frac{1}{N_I}\sum_{i=1}^{N_{I}} (N^{pred}_i - N^{gt}_i)^2}
\end{alignedat}
$}
\end{equation}

where \( N_I \) is the number of test images, \( N_i^{\text{pred}} \) is the predicted count for image \( i \), and \( N_i^{\text{gt}} \) is the corresponding ground-truth count.

\subsection{Implementation Details}
We train the model using the Adam optimizer with a learning rate of $6.25\times10^{-6}$. 
Point-level representations are extracted from the ResNet-50 backbone using feature activations from \texttt{layer1} to \texttt{layer4}. 
All loss balancing weights in Eq.~\ref{eq:loss} are selected via grid search. 
For the encoder, we employ two image–text encoder pairs: CLIP image and text encoders, and a Swin Transformer paired with BERT. More implementation details are provided in the Appendix.
\subsection{Comparison with the State-of-the-Arts}
\input{section/table}


We compare \model with prior counting methods on the FSC147 benchmark, and the results are summarized in Table~\ref{tab:fsc147_comparison}. The baselines span three categories: \textit{few-shot}, \textit{reference-less}, and \textit{zero-shot} methods. Our approach belongs to the zero-shot category.

In the zero-shot setting, where no exemplar information is provided, existing methods typically suffer from higher errors due to the lack of reference guidance. Among the previous approaches, CountGD \cite{NEURIPS2024_57c56985} achieves the runner-up result on the validation set with 12.14 MAE and 47.51 RMSE, while T2ICount \cite{qian2025t2icount} achieves the runner-up result on the test set with 11.76 MAE and 97.86 RMSE. In contrast, \model achieves the best performance on both validation and test sets, obtaining 11.21 MAE / 42.77 RMSE on the validation set and 11.32 MAE / 97.16 RMSE on the test set. These improvements highlight the effectiveness of \model in capturing object-level information without exemplars through structural consistency between density and point representations.

To evaluate cross-dataset generalization, we further test \model on the CARPK and PUCPR+ datasets \cite{hsieh2017drone}. The results are summarized in Table \ref{tab:carpk_puc}. 
Although CountGD reports the lowest error on CARPK, \model ranks second overall and outperforms recent zero-shot counting methods including CLIP-Count~\cite{jiang2023clip}, CounTX~\cite{AminiNaieni23}, and T2ICount~\cite{qian2025t2icount}.
On the PUCPR+ dataset, \model achieves an MAE of 24.27 and RMSE of 28.91, surpassing previous methods such as CountGD \cite{NEURIPS2024_57c56985}. These results demonstrate the strong generalization capability of \model across different datasets and confirm its robustness in zero-shot object counting scenarios.

\subsection{Qualitative Results}
Fig.~\ref{fig:qual} presents qualitative comparisons between \model and existing density based zero-shot counting methods, including VLCounter, CLIP-Count, and CounTX. We visualize the predicted density maps to highlight differences in spatial allocation of density mass. Compared with prior approaches, \model produces more concentrated and accurate density estimates. This improvement is mainly attributed to the proposed instance-level structural constraints. The Per-Instance Mass Conservation encourages balanced density allocation across predicted instances and prevents excessive concentration or dispersion of mass. In addition, the Center-of-Mass Alignment constraint encourages each density component to align with its corresponding predicted point, reducing fragmented or part-biased density assignments.
As a result, \model effectively reduces false positives in challenging scenarios, including self-similar objects (e.g., nail polishes), non-convex object layouts (e.g., camels), crowded scenes (e.g., toilet paper rolls), and distracting backgrounds (e.g., polka dots). In contrast, previous methods often spread density mass into irrelevant regions, leading to inaccurate counting predictions.

\subsection{Ablation Study}

We conduct ablation studies to analyze the contribution of the proposed loss functions and architectural components of \model. The results are summarized in Tables~\ref{tab:ablation_loss} and~\ref{tab:ablation_arch}.

\paragraph{Effect of loss components.}
Table~\ref{tab:ablation_loss} evaluates the impact of each proposed loss term. The full model achieves the best performance with an MAE of \textbf{11.32} and RMSE of \textbf{97.16}. Removing the center-of-mass alignment loss $\mathcal{L}_{align}$ leads to a noticeable degradation (MAE increases from 11.32 to 12.84), indicating its importance in aligning density components with predicted instance centers. Similarly, removing the per-instance mass conservation loss $\mathcal{L}_{mass}$ increases the MAE to 13.05, suggesting that balanced density allocation across predicted instances is critical for accurate counting. Excluding the count consistency loss $\mathcal{L}_{consistent}$ further degrades performance (MAE 14.25), demonstrating the importance of enforcing agreement between density-based and point-based counting estimates. Finally, removing the contrastive loss $\mathcal{L}_{cst}$ leads to the largest performance drop (MAE 14.49), highlighting the role of discriminative point representations in improving localization and counting accuracy.

\paragraph{Effect of architectural components.}
Table~\ref{tab:ablation_arch} studies the influence of different architectural choices. Using only a single decoder significantly degrades performance. Specifically, the density-only model obtains an MAE of 14.12, while the point-only model results in an MAE of 14.54. This demonstrates that the two decoders provide complementary information for counting. Removing the feature enhancer also leads to worse performance (MAE 13.11), indicating that cross-modal feature refinement helps produce more informative spatial representations. We further compare different backbone configurations and observe that replacing the CLIP backbone with the SwinT+BERT encoder improves the MAE from 12.33 to \textbf{11.32}. This confirms that the proposed architecture benefits from richer multi-scale visual features and stronger language representations.

Overall, these results verify that both the proposed loss functions and architectural components contribute significantly to the final performance of \model.

%% file: section/qual.tex
\begin{figure*}[t]
    \centering
    \includegraphics[width=.9\linewidth]{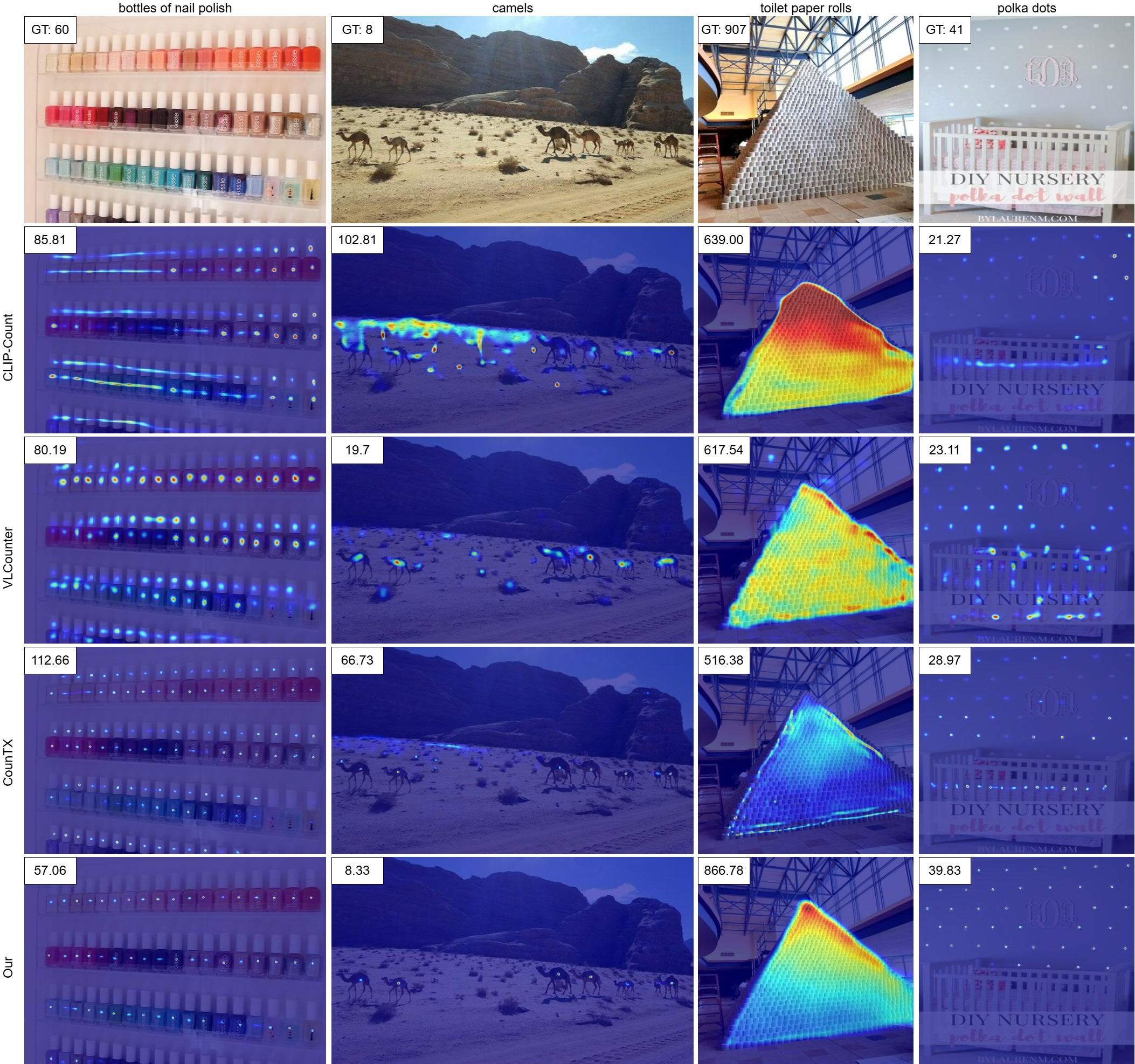}
    \caption{Qualitative comparison between DualCount and density-based zero-shot counting baselines, including VLCounter~\cite{kang2024vlcounter}, CLIP-Count~\cite{jiang2023clip}, and CounTX~\cite{AminiNaieni23}.}
    \label{fig:qual}
\end{figure*}

%% file: section/table.tex
\begin{table*}[t]
\small
\setlength{\tabcolsep}{2pt}
\centering
\caption{\textbf{Quantitative comparison on FSC147.} We report MAE and RMSE under different settings: few-shot, reference-less, and zero-shot. In each setting, the best results are highlighted in \textbf{bold} while the runner-up is highlighted in \textbf{underline}.} 
\resizebox{.9\textwidth}{!}{%
\begin{tabular}{clcccccc}
\toprule
\multirow{2}{*}{\textbf{Scheme}} & \multirow{2}{*}{\textbf{Method}} & \multirow{2}{*}{\textbf{Venue}} & \multirow{2}{*}{\textbf{Shot}} & \multicolumn{2}{c}{\textbf{Val Set}} & \multicolumn{2}{c}{\textbf{Test Set}} \\
\cmidrule(lr){5-6} \cmidrule(lr){7-8}
& & & & MAE & RMSE & MAE & RMSE \\
\midrule

\multirow{8}{*}{Few-shot} 
& FamNet~\cite{ranjan2021learning} & CVPR’21 & 3 & 24.32 & 70.94 & 22.56 & 101.54 \\
& BMNet~\cite{min2022bmnet} & CVPR’22 & 3 & 15.74 & \underline{58.53} & 14.62 & \underline{91.83} \\
& LOCA~\cite{djukic2023low} & ICCV’23 & 3 & \textbf{10.24} & \textbf{32.56} & \textbf{10.97} & \textbf{56.97} \\
& SAM~\cite{shi2023trainingfreeobjectcountingprompts} & WACV’24 & 3 & - & - & 19.95 & 132.16 \\
& PseCo~\cite{li2022pseco} & CVPR’24 & 3 & \underline{15.31} & 68.34 & \underline{13.05} & 112.86 \\
& GMN~\cite{lu2018class} & ACCV’19 & 1 & 29.66 & 89.81 & 26.52 & 124.57 \\
& FamNet~\cite{ranjan2021learning} & CVPR’21 & 1 & 24.32 & 70.94 & 22.56 & 101.54 \\
& BMNet~\cite{min2022bmnet} & CVPR’22 & 1 & 19.06 & 67.95 & 16.71 & 103.31 \\
\midrule

\multirow{3}{*}{Reference-less}
& FamNet~\cite{ranjan2021learning} & CVPR’21 & 0 & 32.15 & 98.75 & 32.27 & 131.46 \\
& LOCA~\cite{djukic2023low} & ICCV’23 & 0 & \textbf{17.43} & \textbf{54.96} & \textbf{16.22} & \textbf{103.96} \\
& RCC~\cite{hobley2022learning} & CVPR’23 & 0 & \underline{17.49} & \underline{58.81} & \underline{17.12} & \underline{104.53} \\
\midrule

\multirow{9}{*}{Zero-shot}
& Patch-selection~\cite{ranjan2022exemplar} & CVPR’23 & 0 & 26.93 & 88.63 & 22.09 & 115.17 \\
& CLIP-Count~\cite{jiang2023clip} & ACM MM’23 & 0 & 18.79 & 61.18 & 17.78 & 106.62 \\
& CounTX~\cite{AminiNaieni23} & BMVC’23 & 0 & 17.10 & 65.61 & 15.88 & 106.29 \\
& VLCounter~\cite{kang2024vlcounter} & AAAI’24 & 0 & 18.06 & 65.13 & 17.05 & 106.16 \\
& PseCo~\cite{li2022pseco} & CVPR’24 & 0 & 23.90 & 100.33 & 16.58 & 129.77 \\
& DAVE~\cite{pelhan2024dave} & CVPR’24 & 0 & 15.48 & 52.57 & 14.90 & 103.42 \\
& VA-Count~\cite{zhu2024zero} & ECCV’24 & 0 & 17.87 & 73.22 & 17.88 & 129.31 \\
& GeCo~\cite{pelhan2024novel} & NeurIPS’24 & 0 & 14.81 & 64.95 & 13.30 & 108.72 \\
& CountGD~\cite{NEURIPS2024_57c56985}  & NeurIPS'24 & 0 & \underline{12.14} & \underline{47.51} & 12.98 & 98.35 \\
& T2ICount~\cite{qian2025t2icount}  & CVPR'25 & 0 & 13.78 & 58.78 & \underline{11.76} & \underline{97.86} \\
& \textbf{\model (CLIP)}  & \textbf{-} & 0 & 14.12 & 54.27 & 12.33 & 99.78 \\
& \textbf{\model (SwinT + BERT)}  & \textbf{-} & \textbf{0} & \textbf{11.21} & \textbf{42.77} & \textbf{11.32} & \textbf{97.16} \\
\bottomrule
\end{tabular}}
\vspace{-1em}
\label{tab:fsc147_comparison}
\end{table*}

\begin{table}[t]
\centering
\setlength{\tabcolsep}{28pt}
\small
\caption{Performance comparison between \model and zero-shot counting baselines on CARPK and PUCPR+.}
\label{tab:carpk_puc}
\resizebox{\columnwidth}{!}{%
\begin{tabular}{@{}l l c c c@{}}
\toprule
Dataset & Method & Venue & MAE $\downarrow$ & RMSE $\downarrow$ \\
\midrule
\multirow{4}{*}{CARPK}
& CLIP-Count \cite{jiang2023clip} & ACM MM'23 & 11.96 & 16.61 \\
& CounTX \cite{AminiNaieni23} & BMVC'23 & 8.13 & 10.87 \\
& T2ICount \cite{qian2025t2icount} & CVPR'25 & 8.61 & 13.47 \\
& CountGD \cite{NEURIPS2024_57c56985} & NeurIPS'24 & \textbf{3.83} & \textbf{5.41} \\
& \textbf{\model (Ours)} & -- & \underline{7.95} & \underline{9.78} \\
\midrule
\multirow{5}{*}{PUCPR+}
& CLIP-Count \cite{jiang2023clip} & ACM MM'23 & 55.82 & 91.23 \\
& CounTX \cite{AminiNaieni23} & BMVC'23 & 75.24 & 113.44 \\
& VLCounter \cite{kang2024vlcounter} & AAAI'24 & 48.94 & 69.08 \\
& CountGD \cite{NEURIPS2024_57c56985} & NeurIPS'24 & 24.31 & 32.16 \\
& \textbf{\model (Ours)} & -- & \textbf{24.27} & \textbf{28.91} \\
\bottomrule
\end{tabular}}
\vspace{-0.8em}
\end{table}

\begin{table}[t]
\centering
\setlength{\tabcolsep}{18pt}
\caption{Ablation study on the proposed loss components. $\checkmark$ indicates the loss is applied during training, while $\times$ indicates it is removed. Test MAE and RMSE are reported for each variant.}
\label{tab:ablation_loss}
\resizebox{\columnwidth}{!}{%
\begin{tabular}{c c c c c c c}
\toprule
$\mathcal{L}_{ot}$ & $\mathcal{L}_{cst}$ & $\mathcal{L}_{consistent}$ & $\mathcal{L}_{mass}$ & $\mathcal{L}_{align}$ & MAE & RMSE \\
\midrule
$\checkmark$ & $\checkmark$ & $\checkmark$ & $\checkmark$ & $\checkmark$ & 11.32 & 97.16 \\
$\checkmark$ & $\checkmark$ & $\checkmark$ & $\checkmark$ & $\times$ & 12.84 & 100.45 \\
$\checkmark$ & $\checkmark$ & $\checkmark$ & $\times$ & $\checkmark$ & 13.05 & 100.98 \\
$\checkmark$ & $\checkmark$ & $\times$ & $\checkmark$ & $\checkmark$ & 14.25 & 102.17 \\
$\checkmark$ & $\times$ & $\checkmark$ & $\checkmark$ & $\checkmark$ & 14.49 & 103.11 \\
\bottomrule
\end{tabular}%
}
\end{table}

\begin{table}[t]
\centering
\setlength{\tabcolsep}{14pt}
\caption{Ablation study on architectural components of \model. We evaluate the impact of the feature enhancer, decoder design, and backbone choice.}
\label{tab:ablation_arch}
\resizebox{\columnwidth}{!}{%
\begin{tabular}{c c c c c c}
\toprule
Enhancer & Density Decoder & Point Decoder & Backbone & MAE $\downarrow$ & RMSE $\downarrow$ \\
\midrule
$\checkmark$ & $\checkmark$ & $\times$ & SwinT+BERT & 14.12 & 103.87 \\
$\checkmark$ & $\times$ & $\checkmark$ & SwinT+BERT & 14.54 & 105.29 \\
$\times$ & $\checkmark$ & $\checkmark$ & SwinT+BERT & 13.11 & 101.08 \\
$\checkmark$ & $\checkmark$ & $\checkmark$ & CLIP & 12.33 & 99.78 \\
$\checkmark$ & $\checkmark$ & $\checkmark$ & SwinT+BERT & \textbf{11.32} & \textbf{97.16} \\
\bottomrule
\end{tabular}}
\end{table}

%% file: section/conclusion.tex
\section{Conclusion}

In conclusion, we presented \model, a zero-shot object counting framework that jointly models density estimation and point-based instance centers. By introducing instance-level structural constraints, per-instance mass conservation and center-of-mass alignment, our method enforces balanced density allocation and geometric consistency, reducing fragmented or biased density predictions.
Extensive experiments on FSC147, CARPK, and PUCPR+ demonstrate that \model achieves state-of-the-art performance in the zero-shot setting while maintaining strong cross-dataset generalization. Ablation studies confirm the contribution of each component, and qualitative results show that \model produces more accurate and concentrated density maps with fewer false positives in challenging scenes. These findings highlight the value of incorporating instance-level structure into density-based counting, and suggest a promising direction for building more robust and generalizable zero-shot counting systems.
More broadly, many computer vision tasks rely on point- or density-map prediction, such as human pose estimation and temporal repetition counting in videos. We believe that the proposed instance-level structural modeling may provide a useful direction for improving density-based prediction beyond object counting.

%% file: main.bib
@String(CVPR  = {IEEE Conf. Comput. Vis. Pattern Recog.})

@String(BMVC  = {Brit. Mach. Vis. Conf.})

@String(AAAI  = {AAAI})

@String(CVPR  = {CVPR})

@String(BMVC  =	{BMVC})

@inproceedings{kang2024vlcounter,
  title={Vlcounter: Text-aware visual representation for zero-shot object counting},
  author={Kang, Seunggu and Moon, WonJun and Kim, Euiyeon and Heo, Jae-Pil},
  booktitle={Proceedings of the AAAI Conference on Artificial Intelligence},
  pages={2714--2722},
  year={2024},
}

@inproceedings{li2022pseco,
  title={Pseco: Pseudo labeling and consistency training for semi-supervised object detection},
  author={Li, Gang and Li, Xiang and Wang, Yujie and Wu, Yichao and Liang, Ding and Zhang, Shanshan},
  booktitle={European Conference on Computer Vision},
  pages={457--472},
  year={2022},
  organization={Springer}
}

@inproceedings{jiang2023clip,
  title={Clip-count: Towards text-guided zero-shot object counting},
  author={Jiang, Ruixiang and Liu, Lingbo and Chen, Changwen},
  booktitle={Proceedings of the 31st ACM International Conference on Multimedia},
  pages={4535--4545},
  year={2023}
}

@inproceedings{xu2023zero,
  title={Zero-shot object counting},
  author={Xu, Jingyi and Le, Hieu and Nguyen, Vu and Ranjan, Viresh and Samaras, Dimitris},
  booktitle={Proceedings of the IEEE/CVF Conference on Computer Vision and Pattern Recognition},
  pages={15548--15557},
  year={2023}
}

@inproceedings{ranjan2021learning,
  title={Learning to count everything},
  author={Ranjan, Viresh and Sharma, Udbhav and Nguyen, Thu and Hoai, Minh},
  booktitle={Proceedings of the IEEE/CVF Conference on Computer Vision and Pattern Recognition},
  pages={3394--3403},
  year={2021}
}

@inproceedings{you2023few,
  title={Few-shot object counting with similarity-aware feature enhancement},
  author={You, Zhiyuan and Yang, Kai and Luo, Wenhan and Lu, Xin and Cui, Lei and Le, Xinyi},
  booktitle={Proceedings of the IEEE/CVF Winter Conference on Applications of Computer Vision},
  pages={6315--6324},
  year={2023}
}

@article{liu2022countr,
  title={Countr: Transformer-based generalised visual counting},
  author={Liu, Chang and Zhong, Yujie and Zisserman, Andrew and Xie, Weidi},
  journal={arXiv preprint arXiv:2208.13721},
  year={2022}
}

@article{kirillov2023segany,
  title={Segment Anything},
  author={Kirillov, Alexander and Mintun, Eric and Ravi, Nikhila and Mao, Hanzi and Rolland, Chloe and Gustafson, Laura and Xiao, Tete and Whitehead, Spencer and Berg, Alexander C. and Lo, Wan-Yen and Doll{\'a}r, Piotr and Girshick, Ross},
  journal={arXiv:2304.02643},
  year={2023}
}

@article{wang2020distribution,
  title={Distribution matching for crowd counting},
  author={Wang, Boyu and Liu, Huidong and Samaras, Dimitris and Nguyen, Minh Hoai},
  journal={Advances in neural information processing systems},
  volume={33},
  pages={1595--1607},
  year={2020}
}

@inproceedings{tyagi2023degpr,
  title={Degpr: Deep guided posterior regularization for multi-class cell detection and counting},
  author={Tyagi, Aayush Kumar and Mohapatra, Chirag and Das, Prasenjit and Makharia, Govind and Mehra, Lalita and AP, Prathosh and others},
  booktitle={Proceedings of the IEEE/CVF Conference on Computer Vision and Pattern Recognition},
  pages={23913--23923},
  year={2023}
}

@inproceedings{ranjan2022exemplar,
  title={Exemplar free class agnostic counting},
  author={Ranjan, Viresh and Nguyen, Minh Hoai},
  booktitle={Proceedings of the Asian Conference on Computer Vision},
  pages={3121--3137},
  year={2022}
}

@inproceedings{min2022bmnet,
  title={Represent, Compare, and Learn: A Similarity-Aware Framework for Class-Agnostic Counting},
  author={Shi, Min and Hao, Lu and Feng, Chen and Liu, Chengxin and Cao, Zhiguo},
  booktitle={Proc. IEEE/CVF Conference on Computer Vision and Pattern Recognition (CVPR)},
  year={2022}
}

@inproceedings{djukic2023low,
  title={A low-shot object counting network with iterative prototype adaptation},
  author={{\DJ}uki{\'c}, Nikola and Luke{\v{z}}i{\v{c}}, Alan and Zavrtanik, Vitjan and Kristan, Matej},
  booktitle={Proceedings of the IEEE/CVF International Conference on Computer Vision},
  pages={18872--18881},
  year={2023}
}

@inproceedings{paiss2023teaching,
  title={Teaching clip to count to ten},
  author={Paiss, Roni and Ephrat, Ariel and Tov, Omer and Zada, Shiran and Mosseri, Inbar and Irani, Michal and Dekel, Tali},
  booktitle={Proceedings of the IEEE/CVF International Conference on Computer Vision},
  pages={3170--3180},
  year={2023}
}

@inproceedings{shi2023trainingfreeobjectcountingprompts,    
    author    = {Shi, Zenglin and Sun, Ying and Zhang, Mengmi},
    title     = {Training-Free Object Counting With Prompts},
    booktitle = {Proceedings of the IEEE/CVF Winter Conference on Applications of Computer Vision (WACV)},
    month     = {January},
    year      = {2024},
    pages     = {323-331}
}

@InProceedings{AminiNaieni23,
  author = "Amini-Naieni, N. and Amini-Naieni, K. and Han, T. and Zisserman, A.",
  title = "Open-world Text-specified Object Counting",
  booktitle = "British Machine Vision Conference",
  year = "2023",
}

@inproceedings{hsieh2017drone,
  title={Drone-based object counting by spatially regularized regional proposal network},
  author={Hsieh, Meng-Ru and Lin, Yen-Liang and Hsu, Winston H},
  booktitle={Proceedings of the IEEE international conference on computer vision},
  pages={4145--4153},
  year={2017}
}

@InProceedings{xc_soict,
author="Cuong, Ngo Xuan
and Mai, Tien-Dung",
editor="Buntine, Wray
and Fjeld, Morten
and Tran, Truyen
and Tran, Minh-Triet
and Huynh Thi Thanh, Binh
and Miyoshi, Takumi",
title="Distribution-Guided Object Counting with Optimal Transport and DINO-Based Density Refinement",
booktitle="Information and Communication Technology",
year="2025",
publisher="Springer Nature Singapore",
address="Singapore",
pages="182--192",
isbn="978-981-96-4282-3"
}

@inproceedings{xc_bmvc,
author    = {Ngo Xuan Cuong},
title     = {Contrastive Point Feature Matching for Open-world Object Counting},
booktitle = {36th British Machine Vision Conference 2025, {BMVC} 2025, Sheffield, UK, November 24-27, 2025},
publisher = {BMVA},
year      = {2025},
url       = {https://bmva-archive.org.uk/bmvc/2025/assets/papers/Paper_1183/paper.pdf}
}

@inproceedings{pelhan2024dave,
  title={Dave-a detect-and-verify paradigm for low-shot counting},
  author={Pelhan, Jer and Zavrtanik, Vitjan and Kristan, Matej and others},
  booktitle={Proceedings of the IEEE/CVF Conference on Computer Vision and Pattern Recognition},
  pages={23293--23302},
  year={2024}
}

@inproceedings{zhu2024zero,
  title={Zero-shot object counting with good exemplars},
  author={Zhu, Huilin and Yuan, Jingling and Yang, Zhengwei and Guo, Yu and Wang, Zheng and Zhong, Xian and He, Shengfeng},
  booktitle={European Conference on Computer Vision},
  pages={368--385},
  year={2024},
  organization={Springer}
}

@article{pelhan2024novel,
  title={A novel unified architecture for low-shot counting by detection and segmentation},
  author={Pelhan, Jer and Lukezic, Alan and Zavrtanik, Vitjan and Kristan, Matej},
  journal={Advances in Neural Information Processing Systems},
  volume={37},
  pages={66260--66282},
  year={2024}
}

@inproceedings{NEURIPS2024_57c56985,
 author = {Amini-Naieni, Niki and Han, Tengda and Zisserman, Andrew},
 booktitle = {Advances in Neural Information Processing Systems},
 editor = {A. Globerson and L. Mackey and D. Belgrave and A. Fan and U. Paquet and J. Tomczak and C. Zhang},
 pages = {48810--48837},
 publisher = {Curran Associates, Inc.},
 title = {CountGD: Multi-Modal Open-World Counting},
 url = {https://proceedings.neurips.cc/paper_files/paper/2024/file/57c56985d9afe89bf78a8264c91071aa-Paper-Conference.pdf},
 volume = {37},
 year = {2024}
}

@inproceedings{qian2025t2icount,
  title={T2icount: Enhancing cross-modal understanding for zero-shot counting},
  author={Qian, Yifei and Guo, Zhongliang and Deng, Bowen and Lei, Chun Tong and Zhao, Shuai and Lau, Chun Pong and Hong, Xiaopeng and Pound, Michael P},
  booktitle={Proceedings of the Computer Vision and Pattern Recognition Conference},
  pages={25336--25345},
  year={2025}
}

@inproceedings{lu2018class,
  title={Class-agnostic counting},
  author={Lu, Erika and Xie, Weidi and Zisserman, Andrew},
  booktitle={Asian conference on computer vision},
  pages={669--684},
  year={2018},
  organization={Springer}
}

@article{hobley2022learning,
  title={Learning to count anything: Reference-less class-agnostic counting with weak supervision},
  author={Hobley, Michael and Prisacariu, Victor},
  journal={arXiv preprint arXiv:2205.10203},
  year={2022}
}

@misc{qian2023counting,
  title={Counting animals in aerial images with a density map estimation model. Ecol. Evol. 13 (4), e9903},
  author={Qian, Y and Humphries, GRW and Trathan, PN and Lowther, A and Donovan, CR},
  year={2023}
}

@article{sindagi2018survey,
  title={A survey of recent advances in cnn-based single image crowd counting and density estimation},
  author={Sindagi, Vishwanath A and Patel, Vishal M},
  journal={Pattern Recognition Letters},
  volume={107},
  pages={3--16},
  year={2018},
  publisher={Elsevier}
}

@inproceedings{guerrero2015extremely,
  title={Extremely overlapping vehicle counting},
  author={Guerrero-G{\'o}mez-Olmedo, Ricardo and Torre-Jim{\'e}nez, Beatriz and L{\'o}pez-Sastre, Roberto and Maldonado-Basc{\'o}n, Saturnino and Onoro-Rubio, Daniel},
  booktitle={Iberian conference on pattern recognition and image analysis},
  pages={423--431},
  year={2015},
  organization={Springer}
}

@inproceedings{ma2019bayesian,
  title={Bayesian loss for crowd count estimation with point supervision},
  author={Ma, Zhiheng and Wei, Xing and Hong, Xiaopeng and Gong, Yihong},
  booktitle={Proceedings of the IEEE/CVF international conference on computer vision},
  pages={6142--6151},
  year={2019}
}
